# *A Portrait of Emotion*: Empowering Self-Expression through AI-Generated Art

Yoon Kyung Lee, Yong-Ha Park, Sowon Hahn

{yoonlee78, peter3061503, swhahn}@snu.ac.kr
Human Factors Psychology Lab, Seoul National University

## Abstract

We investigated the potential and limitations of generative artificial intelligence (AI) in reflecting the authors' cognitive processes through creative expression. The focus is on the AI-generated artwork's ability to understand human intent (alignment) and visually represent emotions based on criteria such as creativity, aesthetic, novelty, amusement, and depth. Results show a preference for images based on the descriptions of the authors' emotions over the main events. We also found that images that overrepresent specific elements or stereotypes negatively impact AI alignment. Our findings suggest that AI could facilitate creativity and the self-expression of emotions. Our research framework with generative AIs can help design AI-based interventions in related fields (e.g., mental health education, therapy, and counseling).

## Introduction

Generative artificial intelligence (AI) is altering the creative process and enabling individuals to express themselves in novel ways, such as creating artwork with simple instructions. This field has the potential to fundamentally change the way we approach the creative process by introducing new perspectives in creative activity using art (Roose, 2022; Slack, 2023). Less emphasis has been paid, however, to the possibilities these models have on *human-ai collaboration* through improving AI alignment (Dafoe et al., 2021; Russel, 2019).

Although it is still debated whether these neural-net-based models truly reflect human intelligence, they have shown great promise in expanding human creativity (Dale, 2021; Mahowald et al., 2023). In-context learning (Brown et al., 2020) of natural language input has led to the creation of text and images that align with human intent, which will help reduce stress and give a broader view of the creative process.

Understanding the problems and possibilities of these technologies can make human-ai collaboration easier; This can also help us figure out how people express their emotions via creative activities, how to represent the semantics of emotion in AI art, and how people can appreciate the process for therapeutic purposes.

Especially creative experiences can help individual mental and physical health by facilitating self-disclosure and emotional understanding of oneself and others. An appropriate criteria and research framework must be established for AI-based creative experiences.

This paper aims to examine how emotions are represented in AI art based on the daily blogs of people, focusing on the potential of text-to-image generative models to expand human creativity and understanding of human emotions.

## Related Work

### Art and Self-Expression

Art has long served as a means of self-expression and for people to identify and experience the meanings and aesthetics of viewing it. Studies have recently compared human-generated and AI-generated art, revealing that it has become increasingly difficult to distinguish between the two (Chamberlain et al., 2018; Gangadharbatla, 2022; Hitsuwari et al., 2023). This distinction becomes even more challenging when the art is generated by the art expert (Chamberlain et al., 2018).

Traditionally, creativity has been evaluated based on a fixed set of criteria, often determined by examining the artwork itself (see Pelowski et al., 2016 for review). An integrative framework for empirical research and theoretical development of the psychological experience of art and aesthetics is relatively recent (Leder & Nadal, 2014).

Previously the aesthetic experience did not regard the role of emotion in the viewer/perceiver of the art; only recently, it became central to the psychological experience of art (Leder et al., 2015). However, the creation process in AI art becomes more collaborative as users generate and modify the art in real-time to express their intended emotional state better. Therefore, revisiting creativity and adjusting assessment criteria to account for the AI's ability to understand and express the user's intended emotional state is crucial.

### Creative Activity and Mental Health

Engaging in creative activities, such as reflective writing or drawing, positively impacts mental and physical health (Esterling et al., 1999; Greenberg et al., 1996). For example, journaling about a stressful or traumatic experience is a commonly adopted task for emotion expression and regulation in therapy (Pennebaker & Beal, 1986; Ullrich & Lutgendorf, 2002) and helps boost positive mood and happiness (Emmons & McCullough, 2003; Lyubomirsky & Layous, 2013).

Reflective writing can be an accessible tool for people of many backgrounds, but other factors can affect individuals' writing experience of expressing emotions. These factors include cultural (Eid & Diener, 2009), individual differences in vocabulary size, personality traits, emotional awareness (e.g., alexithymia; Lumley et al., 2002), and personal preference. Writing requires cognitive efforts, and writing stress can make individuals procrastinate or withdraw from it.

Studies have shown that art can serve as a medium for self-expression and has been found to have the potential to boost well-being as much as reflective writing. Additionally, art is an effective tool for promoting creativity by encouraging individuals to explore new ideas and perspectives (Kim, 2010; MacLeod et al., 2016; Rose & Lonsdale, 2016).

Using digital technology has shown promise in facilitating emotional self-disclosure in a variety of populations, including those who have PTSD (Özkafacı & Eren, 2020), depression (Blomdahl et al., 2018), and stroke (Alex et al., 2021). As a form of creative expression, digital art is growing, and social media platforms offer new ways to express oneself creatively (Loveless, 2003; Lassig, 2012). Traditionally, expressing oneself via digital platforms was viewed as more relevant to younger populations (Hoffman et al., 2016). However, as more people of all ages use social networking, digital art is becoming more diverse in terms of formats (e.g., videos, gifs), platforms (e.g., blogs, newsfeeds), and user groups (experts, non-experts, age).

Therefore, making the process easier to access and alleviating the pressure to write well can be solved by combining AI art generation with reflective writing.

## Generative AIs for Creative Activities

Artificial intelligence technologies that generate natural text and realistic images have experienced a significant increase in development. These generative models have generated photorealistic and aesthetic images (Ramesh et al., 2021), image-to-image translation (Zhu et al., 2017; Wang et al., 2017), image inpainting -reconstructing missing regions within an image (Iizuka et al., 2017), speech synthesis (van den Oord et al., 2016), poetry (Hitsuwari et al., 2023), and creative writing (Yuan et al., 2022) at an above-human level. Multimodal models are particularly adept at creating complex, abstract, or photorealistic works with simple human text instructions, such as "a cute baboon sailing a colorful dinghy at sunset" (Slack, 2023).

The introduction of generative adversarial networks (GANs; Goodfellow et al., 2014 but see 2020), autoregressive models, and variational autoencoders (VAEs; Kingma & Welling, 2013) has shown impressive performances in image generation. GANs consist of a generator and a discriminator, with the generator creating new samples and the discriminator evaluating and providing feedback to improve the output. VAEs also encodes an input image into a low-dimensional latent space and then decode it back to the original image. These multimodal architectures integrate text and images through natural language processing (NLP) and computer vision (CV) techniques and have created new ways of data-driven learning.

In recent years, natural language processing has gained significant attention for its ability to produce human-level products such as creative writing, artwork, audio, and videos (Liu & Chilton, 2022; Wu et al., 2022; Hong et al., 2022). Since the introduction of transfer learning and large language models (LLMs) where there is pre-training of large-scale datasets and model parameters, the performance of in-context learning and matching human alignment has improved dramatically (Brown et al., 2021).

Recent generative models that connect text instruction to images (e.g., 'CLIP'; Radford, 2021) have reached the level where they can generate images based on short text prompts (human instruction), with in-context learning improving their few-shot learning abilities. Advancements in model architecture and the availability of high-quality training data (e.g., image captions, images, videos) from the web have propelled Transformer-based autoregressive approaches like DALL-E (Ramesh et al., 2021) to prominence. Simultaneously, the diffusion-based generative models have progressed, generating more high-resolution, photorealistic images (Ho et al., 2020; Nichol et al., 2022; Rombach et al., 2022; Saharia et al., 2022).

One major limitation of these creative AIs is that they do not possess the capability to share the criteria that humans intuitively use when assessing the creativity of an AI artwork: emotion. Researchers have been exploring ways to align AI with human emotions for reliable human-ai collaboration. However, there is limited research on using these models to promote creativity and mental health by accurately aligning users' intentions and emotions. Efforts have been made such as developing AI music generation software that can produce music that sounds "happier" or "sadder" (CoCoCo; Louie et al., 2020) and creating robot simulators that users can adjust to reflect different moods (e.g., angry-looking; Desai et al., 2019) to improve the interpretability and usefulness of these tools. However, a research gap in understanding how to use AI to promote individuals' creativity and mental health needs to be addressed.

## The Present Study

In this paper, we will present our study on using AI to generate artwork that reflects the mental processes of authors and evaluate the creativity of the AI-generated artwork using criteria from psychology. We will also discuss the implications of our findings for the field of cognitive science and the use of AI in art. We will investigate the use of natural language prompts by users for visual generation in generative frameworks; This approach has received less attention than a text-to-text generation. Here are our contributions in summary:

*Cognitive perspective*: We investigate how semantics from the text are represented by using natural language generation and image generation techniques. Specifically, we investigate how context-learning differs when different focuses are put on: the events and emotions of the author of a narrative story.

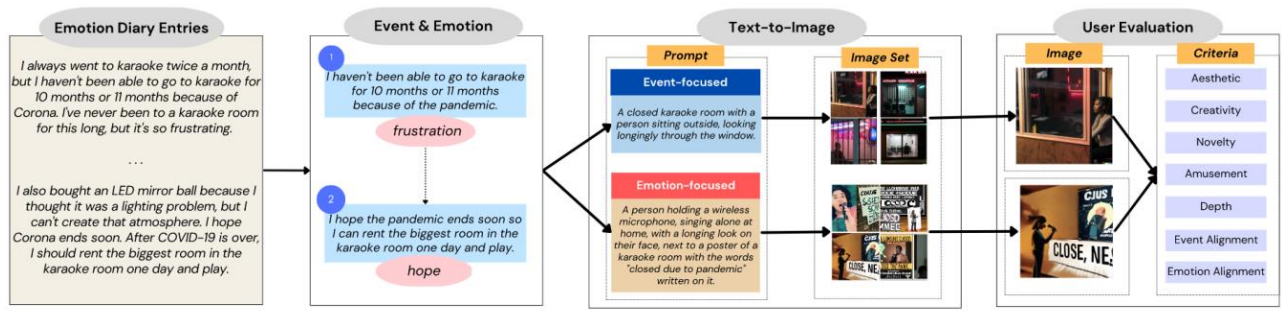

Figure 1: An overall process of generating AI art and evaluation using emotion diary entries. First, the excerpt of the emotional diary entries was summarized in a few sentences by main events and emotions. We also annotated the main emotions detected by a large language model and confirmed them with human evaluators. Each diary was summarized and shortened into two to three main events and emotions (e.g., for the example above, two sentences were selected based on main events. For example, these main events were: 1) "not being able to go to authors' favorite karaoke due to the pandemic," 2) "hope the pandemic ends soon and rent a room from karaoke", and main emotion was "1) Frustration," and "2) Hope"). Then we converted the sentences into main image-captions to generate image sets. Generative models initially generate three to four sample images (candidates) when given a single text prompt (human instruction). Evaluators then selected the final image that best represent the main event and emotions and scored them based on evaluative criteria.

*Human-AI interaction perspective*: We develop an annotation guideline for generating art that accurately reflects users' emotions. Practitioners and researchers in AI-generated art can use it to improve empathy about self and others and create art tailored to the intended audience (e.g., therapeutic art, education, personal use)

*Practical contribution*: Our framework, text prompt, and images are available for cognitive scientists, researchers, practitioners, and evaluators of AI-generated art to consistently and accurately assess the creativity of these works.

## Methods

### Emotion Diary Entries

Figure 1 illustrates overall process. We used diary entries released from the previous study[1], annotated with emotion categories. Each diary entry was self-annotated with one or two emotion categories by diary authors (happy, sad, angry, neutral, disgust, surprise, calm, fear, and others; Ekman, 1992). These diary entries were written during the initial breakout of the pandemic, and people wrote about their daily events within 250 letters. We initially selected 105 diary entries and the final ten entries after controlling the number of entries per emotion, event, and content.

We used the translated version of the diary entries (by expert translators), then summarized the document into a few sentences to annotate specific emotions per sentence and observe the flow of emotion change throughout the story. Neural-net-based models have shown evolved performance in emotional meanings embedded in natural language (Cowen et al., 2021; Brooks et al., 2022) and also from writings (Demszky et al., 2020). We used a neural-net-based large language model, ChatGPT[2], the most recent large language model released by OpenAI[3], to get the sentence-level emotion and confirmed it with two human evaluators.

### Main Event and Emotion Extraction

We extracted three components from the diary entry: 'sentence (from emotion diary)', 'emotion (from the sentence)', and 'image caption (for text-to-image generation)'. At the beginning of the prompt, we always put a short description of a diary author persona (e.g., "AI Persona: You are the author of this diary:"). Defining the persona in the initial step ensured the model interpreted the diary entry from the first-person narration and generated consistent context and style. Otherwise, output sentences would look like different authors wrote them. We, however, did not wish the model to be affected too much by the specific tone or use of language that would reflect the stereotype of demographic profiles (e.g., gender, age), so we kept the persona profile as general as possible.

After setting the persona, we summarized the diary entry into a few sentences. We ensured these summary sentences included changes in the author's emotional state. We, therefore, divided the authors' entire diary entries into one or several sentences that include the main event and emotion.

For image caption, we first applied a prompt to make ChatGPT entirely focus on extracting main events information from the diary and converting it to an image caption that is proper for image generative AI. The prompt of 'image caption focusing on main events from the diary:' was used to make ChatGPT generate an image caption appropriate for text-to-image models. We limited the length of words used for image captioning to prevent overfitting image generation.

---

[1] Detailed information about the original data will be disclosed later in a camera-ready version due to the anonymity rule

[2] https://chat.openai.com/chat

[3] https://openai.com/

When programming a prompt for generating image captions for DALL-E, we explicitly instructed the model to 'focus on both main events and sentiment from the diary.' If these were not emphasized, the model would generate image captions that only capture objective information (e.g., objects, places, people), which would then be irrelevant to our objectives; both main events and authors' emotions were essential components of these generative AIs should attend. Therefore, we designed two conditions: *event-focused* and *emotion-focused* prompts[4].

### AI Art Generation

We used DALL-E 2[5] and StableDiffusion[6] interchangeably to generate images using the same prompt to select the most appropriate images per diary to ensure that its main events and emotions are represented in their drawings. Before generation, we adjusted the format of a few words in sentences analogous to training data (i.e., image captions) for text-to-image models. These changes include removing abstract phrases like "if" and "probably," prepositions, postpositions, and other wordy phrases. Each image caption prompt contained the main event(s) and emotion(s).

We also replaced some words from sentences or prompts, "COVID-19" or "COVID" with "pandemic." This replacement ensured that the model does not give out false information or representations of people affected by the epidemic and respects distributors' ethical policies to avoid spreading potential misinformation regarding the epidemic. We used both automatic and manual approaches to edit these prohibited words (e.g., Corona and COVID-19).

We also limited the sentence length (within 200 to 300 letters) so that one sentence could focus on a single main event and emotion simultaneously. A sentence could imply a combination of multiple emotional categories, which we then limited to a maximum of two based on its saliency.

### User Evaluation

We prepared AI-generated text outputs (emotion detected, prompt for text-to-image model) and the image generated. We asked a group of users to judge AI art with our suggested criteria. We first recruited five evaluators: three graduate students and two undergraduate students with interdisciplinary backgrounds (psychology, cognitive science, engineering). We first checked whether they had previous experience using generative models, ranging from text-to-text to text-to-video. Except for two people who had used

Table 1: Mean score of evaluative criteria per prompt condition

| Criteria | Event-focused | Emotion-focused |
|---|---|---|
| Aesthetic | 2.50 (1.28) | 2.74 (1.23) |
| Creativity | 2.28 (1.18) | 2.78 (1.31) |
| Novelty | 2.42 (1.23) | 2.76 (1.27) |
| Amusement | 2.34 (1.24) | 2.80 (1.26) |
| Depth | 2.34 (1.24) | 2.66 (1.44) |
| Event Alignment | 2.76 (1.17) | 3.04 (1.07) |
| Emotion Alignment | 2.76 (1.30) | 3.30 (1.23) |

*Note*: Mean (Standard Deviation).

ChatGPT, GPT-3, and DALL-E 2 once, they all had yet to experience using these models.

Evaluators were asked to read descriptions of how these images were generated and were provided with background information on the creation process. Each item contained 1) a set of images (three to four generated by the text-to-image models), 2) evaluation criteria (5-item Likert scale), and 3) a text box for free comment. Because we did not want our researchers' judgment of 'the best' image that depicts the story, evaluators were asked to first pick one image from the image set, then proceed to answer the evaluation criteria items. The evaluation criteria included items scores of 1) aesthetic ('how aesthetic the artwork is perceived'), 2) creativity ('how creative the artwork is perceived'), 3) novelty ('how novel the artwork is perceived'), 4) amusement ('how much amusement one would feel from viewing the artwork'), 5) depth ('the depth of meaning the artwork portrays'), 6) event alignment ('how much the image align with the main event of the diary entry'), 7) emotion alignment ('how much the image align with the emotion of the diary entry'). Evaluators were then asked to freely comment about what elements of the artwork made them score for each questionnaire and how the artwork could improve to reflect the main story or emotion from the diary.

### Results

Table 1 shows the mean scores for each criterion per prompt condition ($N_{event} = 10$, $N_{emotion} = 10$). Evaluators rated images generated from emotion-focused prompts as more aesthetic than event-focused prompts ($M_{emotion} = 2.74$ ($SD = 1.23$), $M_{event} = 2.5$ ($SD = 1.28$)). The same tendency appeared from all the other evaluation criteria, including creativity ($M_{emotion}$

---

[4] Specifically, the emotion-focused prompts contain both the main event and emotions (i.e., 'event-and-emotion-focused') of the diary entries. However, because the prompt does have an additional component that the model should attend to (i.e., emotion), we will refer to it as 'emotion-focused' for convenience.

[5] https://openai.com/dall-e-2/

[6] https://huggingface.co/spaces/stabilityai/stable-diffusion

= 2.78 (*SD* = 1.31), $M_{event}$ = 2.28 (*SD* = 1.18), novelty ($M_{emotion}$ = 2.76 (*SD* = 1.27), $M_{event}$ = 2.42 (*SD* = 1.23)), amusement ($M_{emotion}$ = 2.80 (*SD* = 1.26), $M_{event}$ = 2.34 (*SD* = 1.24)), and depth ($M_{emotion}$ = 2.66 (*SD* = 1.44), $M_{event}$ = 2.34 (*SD* = 1.24)).

We conducted a qualitative analysis of the evaluators' comments to find potential common patterns and reasons for scoring each criterion low. Evaluators rated the AI images low in creativity, novelty, and depth. The first factor was *overrepresenting certain physical objects* when the model attended to the meaning of 'objects' more than the overall context. For instance, words like "karaoke," "restaurant," "apartment," and "wearing masks (in the pandemic)" were often used in sentences, and the images generated only emphasized these components.

The second factor was *overrepresenting negative emotions*. For example, although words like "depressed", and "relief", when reading the entire context, the main story was overall positive (e.g., "depressed" -> "relief"). However, the images generated only attended to the emotional words that were either former or negative. This overrepresentation made evaluators demand the model to reason its decision; otherwise, there would be a misrepresentation of the context.

The last factor was *biased against specific ethnic or racial groups*. For example, when asked to draw images of a person traveling in Korea, the generated images included other architecture or landscapes that are more similar to other countries in East Asia. Another example was depicting a person eating dumpling soup, a common food consumed in many countries in the world, but the image focused too much on depicting images of street vendors or signs composed of Chinese characters.

## Discussion

In the present study, we have introduced a novel research framework for utilizing generative AIs to facilitate creative activities for narrating personal stories. We also suggest a new direction in evaluating AI-generated arts where alignment with user intent and context are crucial for creativity and usefulness. As we worked on the research framework, we developed text prompts to generate summarized narratives of emotion diary, emotions per sentence, and text-to-image prompts. We also collected variables that can be further used and provide ground truth for creativity judgment: aesthetic, creative, novelty, amusement, and depth.

### Limitations and Future Studies

The current study leaves several research topics unexplored. First, because this was still at an early stage of development, we did not experiment with more images representing diverse topics and emotions. Future studies should collect more samples to find a potentially significant relationship between emotion alignment in generated AIs and specific emotions or topics. Although our sample size was relatively small ($N_{evaluators}$ = 5), we found potential patterns and relationships among the evaluation criteria that can be further investigated in the follow-up studies.

We learned that in order to create outputs that align with users' mental processes, such as an emotional appraisal of a story, it is crucial to understand how to rephrase natural language into a more efficient programming language. In specific, as we program text prompts for generative models, emphasizing different words between the main event and emotion of the diary entry resulted in different visual representations. Since we humans read a narrative story and evaluate the story and sentiment as a whole (e.g., the overall sentiment of the diary entry), a different approach is needed when putting this perception into explicit descriptions and images.

When evaluating images that were generated by different conditions of prompts (event-focused vs. emotion-focused), evaluators rated the images as aligning more with the authors' intention when the image was made with prompts that represent the emotion of the author (rather than the main event). This pattern is consistent with the previous literature on how non-experts in the art appreciate and interpret art: non-experts are more likely to extract *meaning* out of the artwork they appreciate and relate their personal feelings and memories when interpreting the meaning than experts who may use more domain-specific knowledge (e.g., artistic styles or concepts; Augustin & Leder, 2006; Hager et al., 2012). The follow-up study should explore how these interpretations and a psychological evaluation of AI art differ by expertise.

We found that images that often fail to align with users' intentions. It is suggested to conduct an observational study on how people fix and update each prompt to fine-tune its alignment with their purpose by understanding the author's intention and context.

### Ethical Concerns

Advanced AI models can give the impression that they 'possess' human abilities such as imagination, reasoning, creativity, and emotions. However, it remains controversial whether these neural-network-based models can truly mimic human brain processing. They give the impression that they understand everything, but discretion is advised on how they process and put weight on specific words. For example, when given a prompt that contains specific regions like "Asia," most of the time, the models generate images that depict locations or cultural representations of East Asia. AI models have notoriously been known as biased towards specific regional or racial groups (Caliskan et al., 2017; Kiritchenko & Mohammad, 2018; Manzini et al., 2019). Given this historical background, future research should investigate how to balance and de-bias the weight of specific phrases.

Additionally, people often anthropomorphize human-like traits or actions toward machines (Nass et al., 1994), which can lead to unrealistic expectations. To better utilize AI, it may be more beneficial to treat it as an extended mind rather than a replication of the human mind and investigate its usability and potential to enhance human intelligence and productivity.

## Implications

Lastly, testing its potential in more practical settings such as art education, creativity workshop, or counseling sessions will also be significant. Understanding human intention is critical to leveraging these models for productivity and well-being in creative or therapeutic activities. Therefore, future studies should investigate whether these creative activities help users regulate and express their emotions, including people from diverse backgrounds (from experts to non-experts in art, verbal abilities, and demographic characteristics). In specific, future research could compare whether a more abstract illustration of one's feeling (more like natural human language) or the use of technical language (color code, hue, brightness, RGB value) would result in better artwork that aligns with the users' emotion.

The proposed framework is still in its early stages as new models are increasingly becoming available. However, this approach will help users to engage in reflective writing and express their emotions, which is essential for promoting empathy, and yield a therapeutic effect (Pennebaker, 1997; DasGupta et al., 2004; Garden, 2009; Furman, 2005). We encourage the research community to utilize this dataset and prompt guidelines for beneficial human-AI collaboration.